\documentclass[10pt,twocolumn,letterpaper]{article}

\usepackage[pagenumbers]{wacv} 

\usepackage{scalerel}
\usepackage{tikz}
\usetikzlibrary{svg.path}
\usepackage{multirow}
\usepackage{subcaption}
\usepackage{float}
\usepackage{fixltx2e}

\usepackage{graphicx}

\usepackage[inkscapeformat=png]{svg}

\usepackage{amsmath}
\usepackage{amssymb}
\usepackage{pifont}

\usepackage{comment}
\usepackage{tabularx}
\usepackage{tablefootnote}
\usetikzlibrary{tikzmark}
\usepackage{soul}
\usepackage{booktabs}
\usepackage{arydshln}

\usepackage[pagebackref,breaklinks,colorlinks]{hyperref}

\usepackage[capitalize]{cleveref}
\crefname{section}{Sec.}{Secs.}
\Crefname{section}{Section}{Sections}
\Crefname{table}{Table}{Tables}
\crefname{table}{Tab.}{Tabs.}

\def\wacvPaperID{974} 
\def\confName{WACV}
\def\confYear{2025}

\begin{document}

\title{Shape-biased Texture Agnostic Representations for Improved Textureless and Metallic Object Detection and 6D Pose Estimation}

\author{
Peter Hönig\textsuperscript{1} \quad Stefan Thalhammer\textsuperscript{2} \quad Jean-Baptiste Weibel\textsuperscript{1} \\ Matthias Hirschmanner\textsuperscript{1} \quad Markus Vincze\textsuperscript{1} \\
\textsuperscript{1}TU Wien, \textsuperscript{2}UAS Technikum Vienna\\
}

\maketitle

\begin{abstract}
    Recent advances in machine learning have greatly benefited object detection and 6D pose estimation. 
    However, textureless and metallic objects
    still pose a significant challenge due to few visual cues and the texture bias of CNNs. 
    To address this issue, we propose a strategy for inducing a shape bias to CNN training.
    In particular, by randomizing textures applied to object surfaces during data rendering, we create training data without consistent textural cues.  
    This methodology allows for seamless integration into existing data rendering engines, and results in negligible computational overhead for data rendering and network training.
    Our findings demonstrate that the shape bias we induce via randomized texturing, improves over existing approaches using style transfer.
    We evaluate with three detectors and two pose estimators.
    For the most recent object detector and for pose estimation in general, estimation accuracy improves for textureless and metallic objects.
    Additionally we show that our approach increases the pose estimation accuracy in the presence of image noise and strong illumination changes.
    Code and datasets are publicly available at \url{github.com/hoenigpeter/randomized_texturing}.
\end{abstract}

\section{Introduction}

Object detection and 6D pose estimation are the foundation of robotic manipulation~\cite{bauer2020verefine} and scene understanding~\cite{nie2020total3dunderstanding, thalhammer2023challenges}.
Computational perception and reasoning are generally more challenging for textureless and metallic objects, as they have minimal or no texture cues~\cite{sundermeyer2023bop}.
These objects suffer from inferior detection and pose estimation accuracy due to varying illumination conditions resulting in significantly altered surface appearances.

The standard method for solving this problem is to train with a large amount of synthetic training data, assuming that the data distribution to be expected at runtime is adequately represented~\cite{hodan2019photorealistic,hu2022perspective,labbe2020cosypose,park2019pix2pose,thalhammer2023cope,wang2021gdrnet}.
However, this approach does not consider ambiguities caused by the interaction of the light with the object surface. Consequently, pose estimation approaches designed for such challenging materials focus on retrieving geometrical object representations~\cite{he2023g,yu2023tgf,zeixing2023contourpose,jun2023texturelesspose}.
This strategy is supported by~\cite{geirhos2018imagenettrained}, where the authors demonstrate that CNNs exhibit a texture bias and that ImageNet~\cite{russakovsky2015ImageNet} classification improves when a shape bias is induced.
This fact actually exacerbates the aforementioned phenomenon of increased perception difficulty when processing textureless and metallic objects.

\begin{figure}[t!]
   \centering
    \includegraphics[width=1.0\columnwidth]{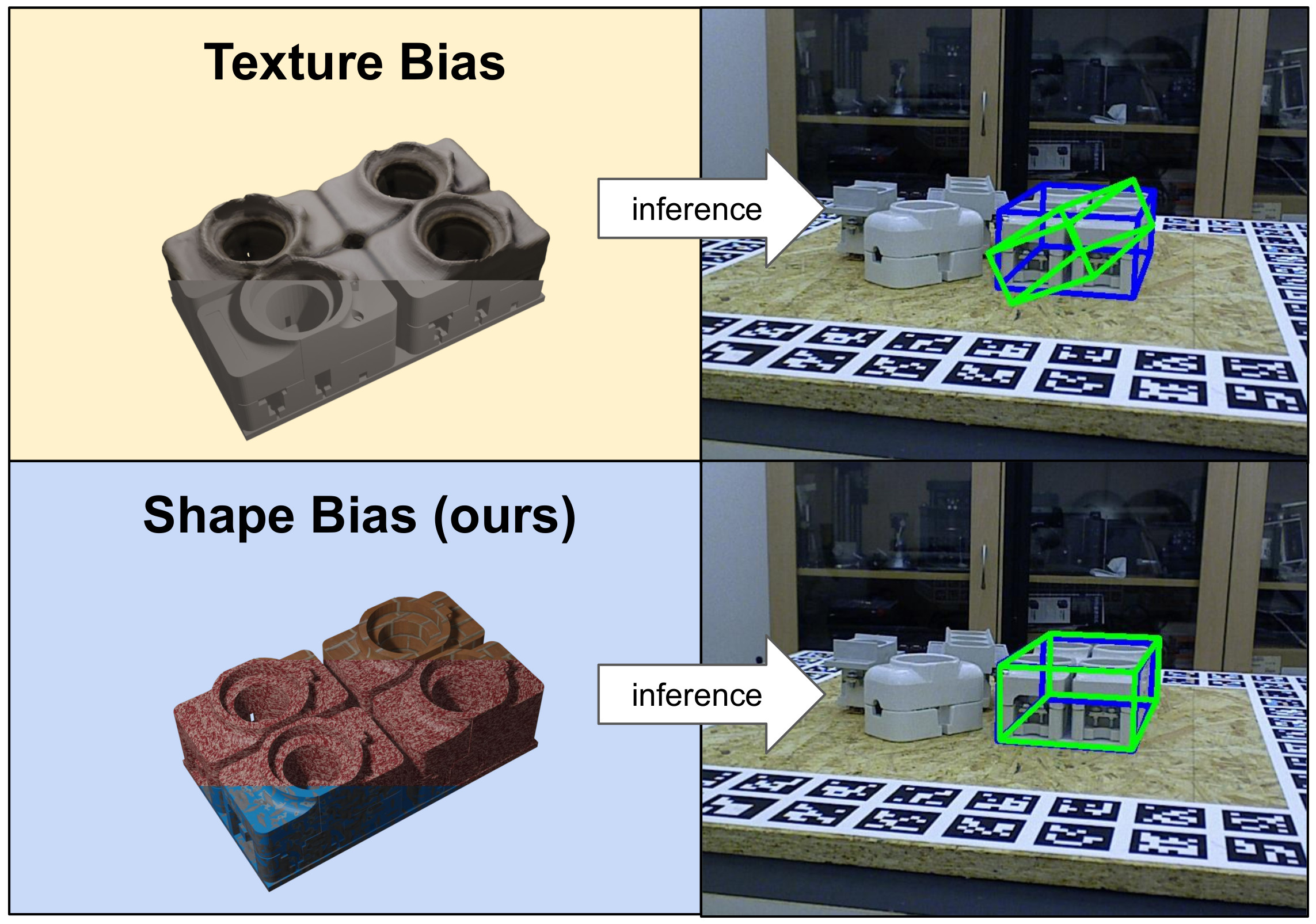}
   \caption{\textbf{Induction of Shape Bias.} 6D pose of a textureless object is visualized with 3D bounding boxes; the ground truth (blue) and estimate (green) of GDR-Net trained with conventional data (top) and when inducing a shape bias (bottom).}
   \label{fig:improvementpic}
\end{figure}

We show that inducing a shape bias during training will be particularly effective for textureless and metallic objects. 
For this purpose we conceptualize texture randomization as a UV-mapping approach.
The surface textures of objects are randomized to render scene-level training data.
This approach offers three key advantages over~\cite{geirhos2018imagenettrained}:
a) UV-mapping of the texture provides geometrically accurate visible cues for observing the object geometry, b) since encoded representations are agnostic to texture, the trained models are more robust with respect to changes in appearance due to illumination, and c) training data is effectively repurposed for multiple vision problems, such as detection, segmentation, and pose estimation.

Experimental results demonstrate that our strategy improves over~\cite{geirhos2018imagenettrained} for the classification of textureless objects. 
Experiments were conducted with three object detectors, YOLOx~\cite{ge2021yolox}, Faster R-CNN~\cite{ren2015faster}, and RetinaNet~\cite{lin2017focal}, and with two pose estimators, GDR-Net~\cite{wang2021gdrnet} and Pix2Pose~\cite{park2019pix2pose}. The experiments provide precise information regarding the improvements and limitations for textureless and metallic objects.
Furthermore, experiments are presented showing enhanced robustness against image perturbations and regarding the object mesh origin.

We summarize our contributions as follows.
\begin{itemize}
    \item We show that the induction of a shape bias to learned representations through the application of our UV-mapping approach improves textureless object classification outperforming style transfer.
    \item The shape bias can be applied to any detection and pose estimation method and we report improvements for three object detection and two pose estimation approaches for textureless and metallic objects. 
    \item Our strategy provides a partial alleviation for the need for online data augmentation. In particular, when recognizing textureless objects with YOLOx, the shape distortion leads to an object recognition accuracy comparable to that achieved by a grid search using the hyperparameters of the online data augmentation.
    \item The induction of the shape bias also robustifies the estimation of the object pose against typical real-world image perturbations, such as noise and significant illumination alterations.
\end{itemize}

The paper is structured as follows: Section~\ref{Related Work} provides an overview of related literature, Section~\ref{Randomized Texturing Approach} introduces the rationale behind the induction of a shape bias, followed by the comprehensive description of the implementation and the experimental setup. Section~\ref{Results} presents the empirical evidence supporting our findings, while Section~\ref{Conclusion} summarizes the observations made and outlines future work.

\section{Related Work}
\label{Related Work}
This section summarizes the state of the art for object detection, pose estimation, and data representation.

\subsection{2D Object Detection and 6D Pose Estimation}
Throughout this paper, the term object detection defines the 2D detection of an object, defined by the class label $c$ and bounding box $\mathbf{b}$.
Pose estimation refers to the instance-level 6D pose estimation of an object, defined by the translation $\mathbf{T}$ and rotation $\mathbf{R}$.

The Benchmark for 6D Object Pose Estimation (BOP)~\cite{sundermeyer2023bop} challenge ranks the best-performing object detection and pose estimation methods, evaluated with standardized metrics and datasets.
Successful object detectors from the last years' challenges include two-stage models such as Mask R-CNN~\cite{he2017mask} and Faster R-CNN, and single-stage ones, e.g. YOLOx.

Pose estimators are either composed of a single stage~\cite{thalhammer2023cope, thalhammer2021pyrapose} or multiple stages~\cite{park2019pix2pose, wang2021gdrnet}, where subsequent stages build upon location priors, provided by arbitrary object detectors.
For each location prior, the subsequent pose estimator performs feature extraction, learns 2D-3D correspondences, and regresses translation and rotation of the object pose.
Successful CNN-based methods~\cite{wang2021gdrnet, park2019pix2pose} from the last years' challenges use multi-stage approaches employing one of the abovementioned object detectors.

\subsection{Data Representation}
State-of-the-art benchmarking datasets for object detection and pose estimation of textureless and metallic objects include TLESS~\cite{hodan2017tless} and ITODD~\cite{drost2017itodd} which consist of physically-based-rendered (PBR) training sets and real-world test sets.
Training with synthetic and evaluating on real-world data leaves a domain gap~\cite{tobin2017domainrandomization}.
Data rendering and model training offer different techniques for bridging this domain gap.
During rendering, domain randomization~\cite{tobin2017domainrandomization} can be applied, which includes the variation of backgrounds, camera views, object poses, and lighting.
After rendering, during the training procedure data augmentation can be applied online, either with pre-trained augmentations \cite{zoph2020learning, behpour2019ada}, style-transfer \cite{geirhos2018generalisation} or handcrafted combinations of varying image perturbation types.

The domain gap is also influenced by the quality of the available object geometry and texture.
Depending on the target domain, meshes for the datasets are either provided via CAD models or reconstructions~\cite{mildenhall2021nerf,yang2018tsdf}.
TLESS and ITODD, for example, feature plastic and metallic objects for industrial purposes, where CAD models are usually available.
CAD models are geometrically accurate but often lack texture information, while reconstructions have added texture information but less geometric accuracy due to reconstruction noise.

While existing methods for bridging the domain gap focus on textured objects with distinct visual cues, less attention has been paid to textureless and metallic objects.
Since lighting, reflections and shadows alter the appearance of these objects, they do not have continuous textures in the real-world domain.
We hypothesize that texture should be treated as an unknown and object detection as well as pose estimation models should be biased toward shape.
We therefore want to explore the randomization of object textures, to force CNNs to learn exclusively from object geometries from 2D RGB images only.


\begin{figure*}[h!]
   \centering
    \includegraphics[width=2\columnwidth]{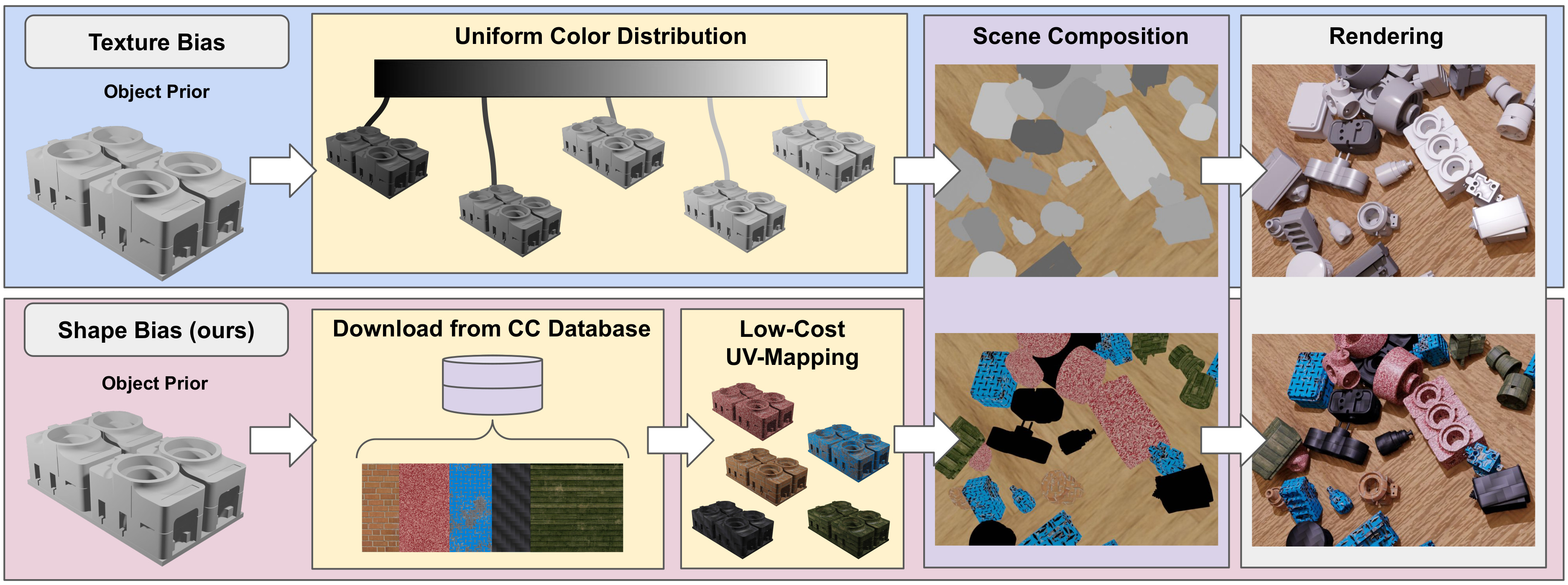}
   \caption{\textbf{Synthetic Data Generation with Randomized Texturing Pipeline.} Instead of sampling color values from a uniform color distribution we download textures from a creative-commons database and perform low-cost UV-mapping, a method that we call texture randomization. While the figure shows the scene composition and rendering steps with five exemplary textures, the texture randomization method can be used with an arbitrary number of texture files.}
   \label{fig:mesh_to_pose_pipeline}
\end{figure*}

\section{Induction of Shape Bias}
\label{Randomized Texturing Approach}

This section presents the rationale of biasing representation learning toward shape.
The authors of~\cite{geirhos2018imagenettrained} present evidence that CNNs exhibit a bias toward texture.
By employing style transfer with AdaIN~\cite{gatys2016image} the learned representations of the network are shape biased, thereby enhancing classification on ImageNet~\cite{russakovsky2015ImageNet}.
Particularly for textureless and metallic objects, where the texture is sparse the phenomenon of texture bias is disadvantageous.
This has also been observed in the context of object pose estimation for textureless and metallic objects.
As a result, these objects are typically handled by estimating their shape, for example by estimating edges and silhouettes~\cite{he2023g,yu2023tgf,zeixing2023contourpose,jun2023texturelesspose}.

The capacity of networks to generalize from synthetic to real object primitives with random color has been demonstrated by~\cite{tobin2017domainrandomization}. 
The authors consider various scene composition parameters, including background texture and camera, object, and light source poses as random variables.
Similarly~\cite{denninger2019blenderproc} observed that replacing the scene background with textures is an effective method for foreground-background separation in various vision problems.
This strategy is considered the standard for creating training data for object pose estimation.
We hypothesize that the application of random textures to objects themselves results in encoding the object shape in feature spaces, a process analog to that observed by~\cite{geirhos2018imagenettrained}. 
This hypothesis is tested through the formulation of a learning strategy that biases estimators toward shape through rendering, thus:
\begin{enumerate}
    \item{\textbf{Encoding geometrically accurate cues.} Localization tasks, such as object detection and pose estimation, are susceptible to geometry and illumination effects such as shadows~\cite{kaskman2019homebreweddb}. Rendering allows us to accurately model and control the distribution of such effects.}
    \item{\textbf{Being robust against strong surface appearance changes.} Metallic objects present a particular challenging case as their appearance is strongly influenced by the nature of the incident light. By learning shape representations, it is possible to achieve a certain degree of agnosticism with regard to surface effects.}
    \item{\textbf{Obtaining training data for different problems.} Although the authors of~\cite{geirhos2018imagenettrained} demonstrate that style transfer is a viable strategy for object classification, we present a strategy for inducing a shape bias for classification, detection and pose estimation. While not yet tested, our strategy can theoretically be applied to the same set of vision problems as that of~\cite{denninger2019blenderproc}.}
    \item{\textbf{Being more universally applicable.} A significant limitation of instance-level approaches is that the trained models lack the capacity to generalize effectively to changes of the object. While this remains unsolved, our approach offers a degree of mitigation. As our trained models are color and texture agnostic, we are able to accommodate instances where the object color is unknown.}
\end{enumerate}

Next, we present the experimental setup and implementation detail to validate our hypothesis.

\begin{table}[t!]
\centering
\caption{\textbf{Domain Randomization Overview.} Parameters and functions denoting the rendering specifications for the TLESS and ITODD objects}
\label{table:domainrandomization}
\begin{tabular}{ccc}
\toprule
\textbf{Method} & \textbf{Parameter} & \textbf{Function}\\
\midrule
\multirow{3}{*}{\shortstack{Texture \\Bias}} & Specularity & $S \sim \mathcal{U}(0, 1)$ \\
& Roughness & $R \sim \mathcal{U}(0, 1)$ \\
& Color & $(R, G, B) \sim \mathcal{U}(0.1, 0.9)$ \\
\midrule
\multirow{3}{*}{\shortstack{Shape \\Bias (ours)}} & Specularity & $S \sim \mathcal{U}(0, 1)$ \\
& Roughness & $R \sim \mathcal{U}(0, 1)$ \\
& Color & $(R, G, B) = T_{n\sim\mathcal{U}(1, 1226)}$ \\
\bottomrule
\end{tabular}
\end{table}

\section{Data Rendering for Inducing a Shape Bias}
\label{sec:rendering}
We employ physically based rendering (PBR) to accurately model and project the object geometry to the image space and cast realistic shadows to induce a shape bias in estimators.
This concept has been demonstrated to be beneficial for detection and pose estimation under the synthetic-to-real setting~\cite{hodan2019photorealistic,sundermeyer2023bop}.

The standard tool for rendering data for object pose estimation is BlenderProc~\cite{denninger2019blenderproc}.
The procedure is delineated in Figure~\ref{fig:mesh_to_pose_pipeline}.
Conventional approaches either assume the availability of a texture; obtained through object reconstruction or by a texture file, or a color prior for textureless and metallic objects.
Instead of relying on these strategies, we hypothesize that they are detrimental to the representations learned during training.
Therefore, we uniformly sample from a set of $n$ textures and apply those via UV-mapping for PBR rendering of the training data.

Scene composition and rendering are executed using~\cite{denninger2019blenderproc} with the standard configuration of BOP~\cite{sundermeyer2023bop}.
Table~\ref{table:domainrandomization} illustrates the list of object material parameters and their ranges utilized for scene composition.
The specular and roughness values are assigned identical values to those used within the standard configuration\footnote{\url{https://github.com/DLR-RM/BlenderProc/tree/main/examples/datasets/bop_challenge}}. However, to induce a shape bias, color randomization is replaced with our own texture randomization strategy.
Let $T=\{T_1 , T_2 ,\dots, T_n\}$ be a set of textures and $M=\{M_1 , M_2,\dots, M_k\}$ a set of objects.
The random texture is drawn from $T$, with $n$ being a hyperparameter denoting the number of textures.
The training dataset for $M$ is generated according to the number of scenes $N=\{N_1 , N_2,\dots, N_j\}$ and that for the rendered views per scene.
For each sampled scene $O_{m}$ and object of $M$ the texture is drawn from a uniform distribution $\mathcal{U}_{1,n}$ over $T$, for each sampled object of $M_{k}$. 
For every scene $O_{m}$, $l$ images are rendered with randomized object and camera poses, specular, roughness, and color values or textures.

According to the standard configuration of~\cite{sundermeyer2023bop} the obtained set of $I$ is $50,000$, obtained from $l=25$ images for each of the $m=2000$ scenes in $O$.
Textures are taken from a Creative Commons online texture database (cc-textures)\footnote{\url{https://cc0-textures.com/}}.
The number of random textures is $n=1226$. 
This number represents the complete set of available opaque textures in the database and has negligible computational overhead in comparison to a smaller number.
As test sets, we use the ones provided by BOP.

\section{Experiments}
\label{Results}

This section presents the empirical validation of our contributions.
The experiments on object classification demonstrate the advantage of biasing CNNs toward shape with our approach over that of~\cite{geirhos2018generalisation}.
Following this, an analysis regarding the advantages and limitations of our approach for object detection and pose estimation of textureless and metallic objects is presented.
Subsequently, the robustness to typical real-world image perturbations is demonstrated.
Ultimately, a series of ablations are presented to ascertain the significance of the number of textures utilized for rendering, to effectively integrate the presented shape bias with online data augmentation, and to investigate the influence of the mesh origin, texture, and color.

\subsection{Experimental Setup}
The following paragraphs present the set of methods and the datasets utilized for the validation our hypothesis, and the metrics employed for comparison.

\subsubsection{Methods}
The classification experiments are conducted using ResNet50~\cite{he2016deep}.
The object detection results are provided with Faster R-CNN \cite{ren2015faster}, RetinaNet\cite{lin2017focal}, and YOLOx \cite{ge2021yolox}.
These three approaches were considered the state of the art at different points in time over the last few years.
The same applies to our object pose estimation experiments where results are presented with GDR-Net~\cite{wang2021gdrnet} and Pix2Pose~\cite{park2019pix2pose}.

\begin{table*}[t!]
  \centering
  \caption{\textbf{Object Detection.} mAP\textsuperscript{0.50:0.95} scores for Faster R-CNN~\cite{ren2015faster}, RetinaNet~\cite{lin2017focal}, and YOLOx~\cite{ge2021yolox}  trained on TLESS and ITODD. Results are provided for learning a texture and a shape bias, with and without tuned online data augmentations.}
  \label{tab:detection-data}
  \begin{tabular}{ccc|cc|cc|cc}
    \toprule
    \multirow{2}{*}{\textbf{Dataset}} & \multirow{2}{*}{\textbf{{\shortstack{Data \\Augmentation}}}} & & \multicolumn{2}{c}{\textbf{Faster R-CNN}} & \multicolumn{2}{c}{\textbf{RetinaNet}} & \multicolumn{2}{c}{\textbf{YOLOx}} \\
    \cmidrule{4-5}
    \cmidrule{6-7}
    \cmidrule{8-9}
    & & Bias: & texture & \textbf{shape (ours)} & texture & \textbf{shape (ours)} & texture & \textbf{shape (ours)} \\\hline
    \multirow{2}{*}{TLESS} & default & & 20.20 & \textbf{70.10} & 23.10 & \textbf{69.50} & 52.60 & \textbf{79.70} \\
    & tuned & & \textbf{80.50} & 77.90 & \textbf{78.40} & 74.10 & 80.20 & \textbf{80.60} \\ \hline
    \multirow{2}{*}{ITODD} & default & & 21.10 & \textbf{23.50} & \textbf{15.00} & 13.90 & 25.70 & \textbf{41.20} \\
    & tuned & & 46.20 & \textbf{46.60} & 41.50 & \textbf{44.20} & 53.10 & \textbf{56.30} \\ \bottomrule
  \end{tabular}
\end{table*}
\subsubsection{Datasets}
In order to validate our hypothesis that training models for a shape bias is particularly beneficial for textureless and metallic objects, evaluations on TLESS~\cite{hodan2017tless} and ITODD~\cite{drost2017itodd} are presented.
The TLESS dataset comprises $30$ highly symmetrical textureless industrial objects.
The ITODD dataset contains $28$ metallic industrial objects.
In both cases, the BOP test sets are employed for the evaluation, and the results for the texture bias baselines are obtained using the PBR training sets of the BOP~\cite{sundermeyer2023bop}.
These were generated using the rendering hyperparameters outlined in Section~\ref{sec:rendering}.
The ground truth of the ITODD test set is not available to prohibit users from optimizing their data augmentation hyperparameters for the test data.
Thus, evaluations can only be performed with the online evaluation tool of the BOP, which only provides metrics for object detection and pose estimation.
Consequently, experiments for classification and the ablations are done using TLESS.

\subsubsection{Metrics}
For the classification experiments the top-1 and the top-5 accuracy in percentage are reported.
The COCO evaluation metric~\cite{lin2014microsoft}, mean average precision (mAP), is reported for object detection. The intersection over union (IoU) range emploed is $0.5:0.95$.
For pose estimation, we present the Average Recall (AR)~\cite{hodan2018bop}, the combination of the visible surface discrepancy (VSD)~\cite{hodan2018bop}, maximum symmetry-aware surface distance (MSSD)~\cite{hodan2020bop}, and the maximum symmetry-aware projection distance (MSPD)~\cite{hodan2020bop}.

\subsection{UV-Mapping versus Style Transfer}

\begin{table}[h!]
\centering
\caption{\textbf{UV-Mapping versus Style Transfer.} Comparing top-1 and top-5 accuracy (Acc.) of style transfer for shape bias induction~\cite{geirhos2018imagenettrained} and our approach on the TLESS dataset.}
\label{table:uvmapping_style_transfer}
\begin{tabular}{c|cc}
\toprule
\textbf{Bias} & \textbf{Top-1 Acc. (\%)} & \textbf{Top-5 Acc. (\%)}\\
\midrule
texture & 63.22 & 86.27 \\
shape\cite{geirhos2018imagenettrained} & 75.88 & 91.16 \\
shape (ours) & \textbf{87.27} & \textbf{97.37} \\
\bottomrule
\end{tabular}
\end{table}

We conduct an experiment to verify our hypothesis that our UV-mapping approach is improving over style transfer for learning a shape bias.
Table~\ref{table:uvmapping_style_transfer} presents results for ResNet50~\cite{he2016deep} trained for object classification on TLESS. 
For classification, all object instances of the test set are cropped using the ground truth bounding box with a scaling factor of $1.0$.
This value correlates with the cropping ratio of ImageNet, which~\cite{geirhos2018imagenettrained} has been designed for.
Our experiments show that learning classification with a texture bias, using uniform color sampling, results in the worst top-1 and top-5 accuracy. 
Using AdaIN~\cite{huang2017arbitrary} for learning a shape bias with style transfer, as done by~\cite{geirhos2018imagenettrained} improves over the texture bias variant.
However, inducing a shape bias through our UV-mapping approach results in a $15.01\%$ relative top-1 and a $6.81\%$ relative top-5 accuracy improvement over~\cite{geirhos2018generalisation}.

\subsection{Object Detection}
Experiments using Faster R-CNN, RetinaNet, and YOLOx for object detection on TLESS and ITODD are conducted to identify the advantages and limitations of our method for shape bias induction.

For Faster R-CNN and RetinaNet the Image-Net pre-trained ResNet50 backbone is used.
For YOLOx the l-variant is used.
All three object detectors are trained for 30 epochs, since validation accuracy converges in all cases until then.
The remaining training parameters, are all identical to the default settings of MMDetection\footnote{\url{https://github.com/open-mmlab/mmdetection}}, which are true to the original papers.

All experiments are conducted using the standard configuration for online data augmentation, the data augmentation of the winner of the BOP $2022$~\cite{liu2022gdrnpp_bop} when learning a texture bias.
Tuned data augmentation is used to maximize the detection accuracy at the superposition with shape bias.
More detail on the superposition of our shape bias and data augmentation is provided in Section~\ref{sec:exp_dataaug}.

\begin{figure}[t!]
    \centering
        \includegraphics[width=0.8\linewidth]{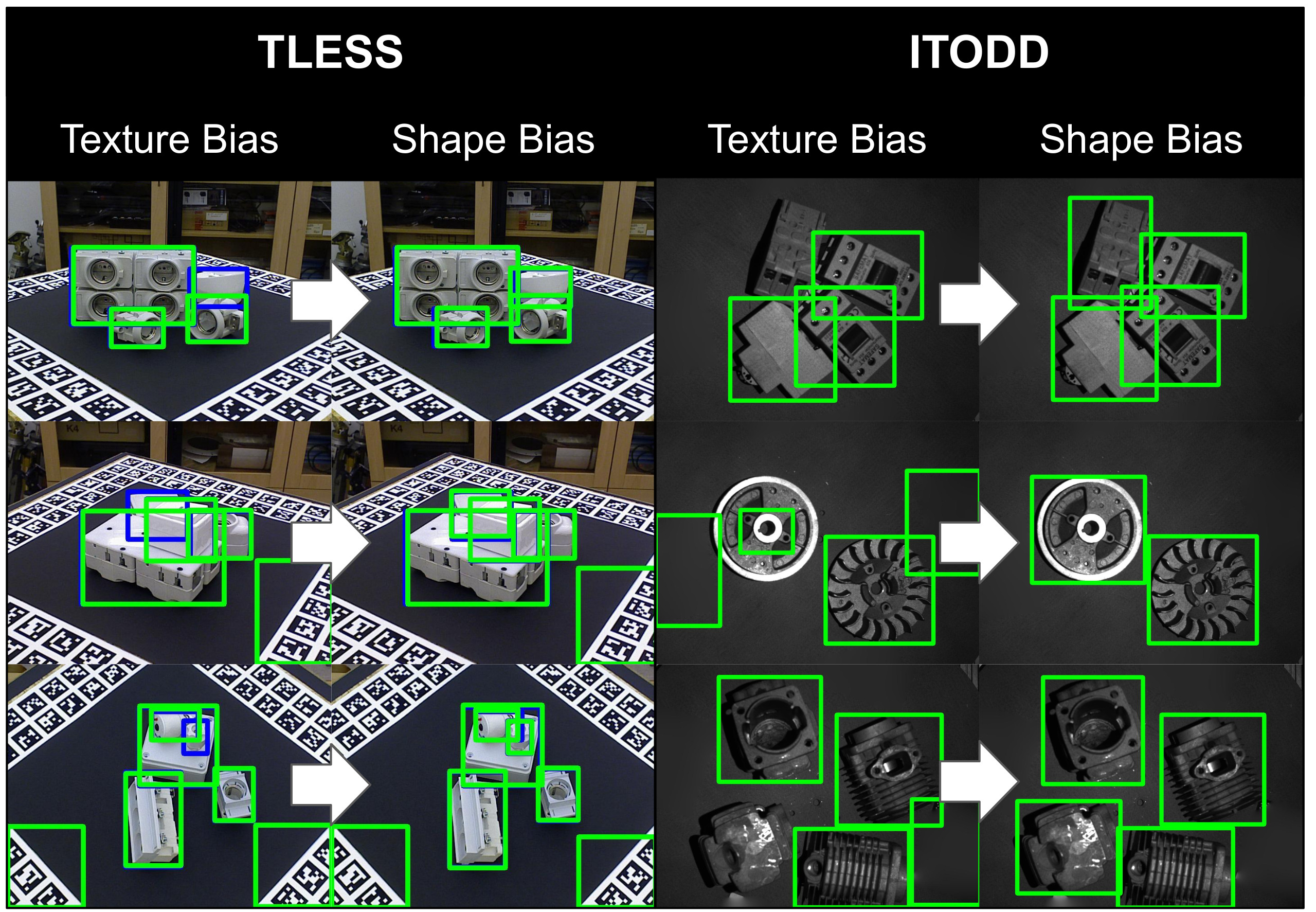}
    \caption{\textbf{Detection Example.} Examplary test images from TLESS and ITODD, with ground truth (blue) and predicted (green) bounding boxes using YOLOx. YOLOx yields more accurate bounding boxes, better recall and precision rates, when inducing a shape bias; detection and IoU thresholds are both set to $0.5$.}
    \label{figure:detection_examples}
\end{figure}
\begin{table*}[t!]
  \centering
  \caption{\textbf{Pose estimation performance.} Pose estimation of GDR-Net and Pix2Pose on TLESS and ITODD. Compared is the AR training with object color priors (default) and for texture agnosticity (none), for object detection (O.D.) and pose estimation (P.E.); YOLOx is used for object detection.}
  \label{tab:tless-data}
  \begin{tabular}{ccc|cccc|cccc}
    \toprule
    \multirow{2}{*}{\textbf{Dataset}} & \multicolumn{2}{c}{\textbf{Bias}} &
    \multicolumn{4}{c}{\textbf{GDR-Net}} & 
    \multicolumn{4}{c}{\textbf{Pix2Pose}}  \\
    \cmidrule{2-7}
    \cmidrule{8-11}
    & O.D. & P.E. & AR\textsubscript{MSPD} & AR\textsubscript{MSSD} & AR\textsubscript{VSD} & \textbf{AR} &
    AR\textsubscript{MSPD} & AR\textsubscript{MSSD} & AR\textsubscript{VSD} & \textbf{AR} \\
    \midrule 
    \multirow{3}{*}{\shortstack{TLESS}} & texture & texture & 72.66 & 50.52 & 45.70 & 56.29 & 68.35 & 37.85 & 33.84 & 46.68 \\
    & texture & \textbf{shape (ours)} & 73.54 & 52.94 & 48.09 & 58.19 & 67.71 & 38.25 & 34.10 & 46.69 \\
    \cdashline{2-11}
    & \textbf{shape (ours)} & \textbf{shape (ours)} & \textbf{73.73} & \textbf{53.05} & \textbf{48.17} & \textbf{58.32} & \textbf{71.22} & \textbf{41.86} & \textbf{37.30} & \textbf{50.13}\\
    \midrule 
    \multirow{2}{*}{\shortstack{ITODD}} & texture & texture & 14.30 & 10.10 & 7.80 & 10.70 & 22.50 & 8.00 & 8.00 & 12.90\\
    & texture & \textbf{shape (ours)} & 14.30 & 10.10 & 7.80 & 10.70 & 22.70 & 10.10 & 10.30 & 14.30\\
    \cdashline{2-11}
    & \textbf{shape (ours)} & \textbf{shape (ours)} & \textbf{15.30} & \textbf{10.20} & \textbf{7.90} & \textbf{11.10} & \textbf{23.10} & \textbf{11.00} & \textbf{11.70} & \textbf{15.30}\\
    \bottomrule
  \end{tabular}
\end{table*}

Object detection results are presented in Table \ref{tab:detection-data}.
For the default two-staged Faster R-CNN and the single-staged RetinaNet our shape bias improves by $247.03$ and $200.87$ relative percent on TLESS, respectively.
On ITODD the improvement in mAP for Faster R-CNN is $11.37\%$.
However, the mAP of RetinaNet decreases by $7.91\%$ when learning a shape bias for object detection on ITODD.
Similarly, when using tuned data augmentations for TLESS our shape prior reduces the mAP by $3.34\%$ for Faster R-CNN and $5.80\%$ for RetinaNet.
On ITODD the relative mAP improvement is $0.87\%$ for Faster R-CNN and $6.51\%$ for RetinaNet.  

On the more recent YOLOx model, which is arguably the current standard for object detection, our shape bias improves the mAP for TLESS and ITODD using the default and the tuned data augmentation version.
Comparing the default versions, our shape bias improves by a relative percentage of $51.52$ for TLESS.  
The improvements on TLESS using a superposition of our shape bias and tuned online data augmentations are negligible to that of using the online data augmentations of~\cite{liu2022gdrnpp_bop} for training for a texture bias. 
However, on the metallic objects of ITODD our relative improvement is $60.31\%$, using the default, and $6.03\%$ using the tuned training configuration.
Interestingly, on TLESS using our method for learning a shape bias with all tested detectors is almost on par with using tuned data augmentations.
As such, a valid alternative when detecting textureless objects.
Metallic object detection on ITODD is more challenging, yet reaching the maximum mAP requires all tested detectors to learn a shape bias.
The experiments demonstrates that YOLOx exhibits enhanced accuracy in comparison to the other two detectors.
We conclude that approaches with higher learning capacities 
benefit more from the induced shape bias.

Figure \ref{figure:detection_examples} shows example images of TLESS and ITODD with bounding boxes predicted by YOLOx, with texture and shape bias.
On TLESS the bounding box corner accuracy and the true positive rate are similar, yet the false negative rate is lower for the shape bias.
On ITODD, the shape-biased YOLOx shows improved accuracy in bounding box corners, higher true positive rate, and lower false negative rate.

In conclusion, inducing object detectors with a shape bias not only improves object detection in the majority of the cases, but also presents a viable alternative to color augmentation tuning for such textureless objects as those of TLESS. For the industrial metallic objects of ITODD the combination of tuned online color augmentation and geometry bias is consistently demonstrated to be the optimal approach.

\subsection{Object Pose Estimation}

This section demonstrates the influence on object pose estimation of textureless and metallic objects when training for a shape bias.
Results are presented for GDR-Net and Pix2Pose, integrated with the detections of YOLOx.
The default training protocols as reported in~\cite{wang2021gdrnet} and~\cite{park2019pix2pose} are employed, with GDR-Net using ResNet34 and Pix2Pose using ResNet50 as backbones.
In both cases, one pose estimator is trained for each object. 
Table~\ref{tab:tless-data} shows that our shape-biased GDR-Net and Pix2Pose improve in AR using the texture-biased version of YOLOx for object detection.
While the improvement with Pix2Pose is negligible, the improvement of GDR-Net is $3.38\%$.
Further integrating the shape-biased YOLOx for detection enhances improvements to $3.61\%$ for GDR-Net, and $7.39\%$ for Pix2Pose.
On ITODD, GDR-Net results in the same AR with texture and with shape bias. 
Integrating the detections of the shape-biased YOLOx improves AR by $3.74\%$.
Using the texture-biased YOLOx, Pix2Pose improves by $10.85\%$.
Integrating the detections of the shape-biased YOLOx the AR improvement is further enhanced to $18.60\%$.
Pix2Pose shows a greater degree of accuracy improvement than GDR-Net.
Analogous to our observations on object detection, Pix2Pose' results improve more due to the higher capacity of the feature extractor. 

\begin{figure}[t!]
   \centering
    \includegraphics[width=0.8\columnwidth]{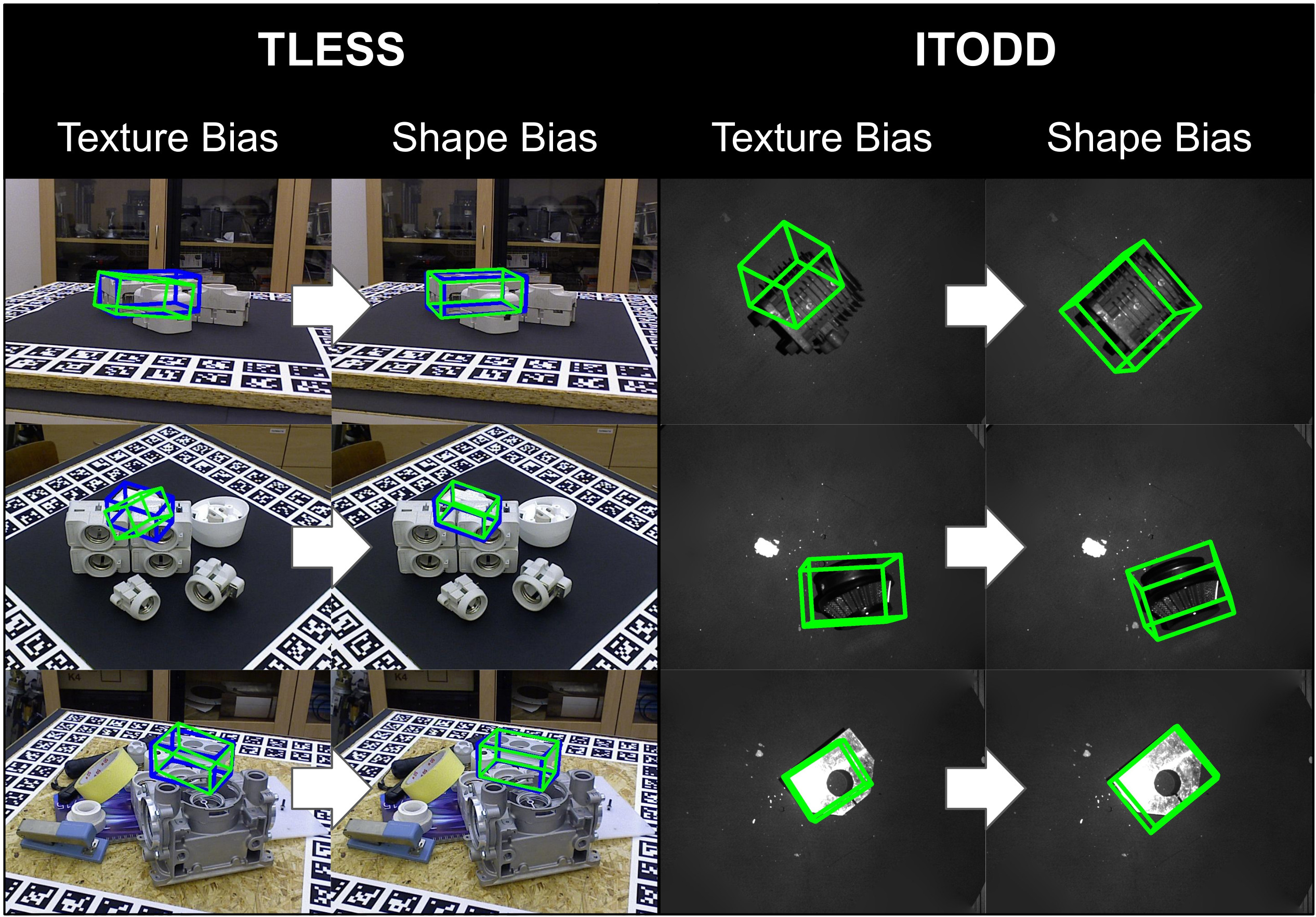}
   \caption{\textbf{Pose Estimation Examples.} Example images from the TLESS and ITODD datasets, showcasing the increased performance for occluded, dark and reflective object pose estimation; GDR-Net for pose estimation and YOLOx without color prior for detection; no detection ground truth for ITODD available.}
   \label{fig:pose_estimation_examples}
\end{figure}

Figure~\ref{fig:pose_estimation_examples} shows qualitative pose estimation results using YOLOx as detector and GDR-Net for pose estimation.
On both, TLESS and ITODD, the alignment of the 3D bounding box, re-projected using the estimated pose, with the object improves.
For TLESS the ground truth is additionally visualized in blue.

In conclusion, biasing representations learned by pose estimators for shape is either on par, or improves accuracy as compared to the standard, texture-biased versions.
Integrating shape-biased object detectors with pose estimators consistently results in pose estimation accuracy improvement.

\subsection{Robustness to Image Perturbations}

\begin{figure*}[t!]
   \centering
    \includegraphics[width=2\columnwidth]{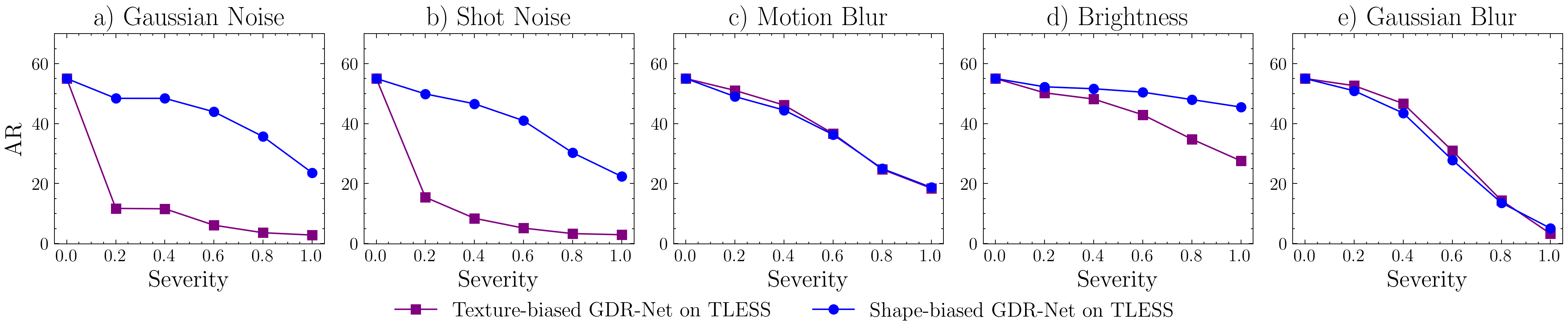}
   \caption{\textbf{Shape Bias Robustness to Image Perturbations} Influence of various image perturbations on AR scores with varying severity; GDR-Net with one pose estimation model trained per object; ground truth detections used for pose estimation}
   \label{fig:perturbations}
\end{figure*}

To assess the robustness of pose estimators trained for shape bias, experiments with different common image perturbations are conducted. 
The perturbations are applied to the test images of TLESS, using ground-truth detections.
Poses are estimated using GDR-Net.
Figure~\ref{fig:perturbations} demonstrates that the shape bias yields higher robustness to Gaussian and Shot image noise.
Pose estimates remain highly accurate even in the presence of considerable noise.
In general, shape bias has little to no influence on the pose estimation accuracy when image blur, such as motion and Gaussian blur, is present.
However, the presented shape bias increases robustness in the regime of high brightness changes, which is potentially useful for cases such as pose estimation in space~\cite{hu2021wide}.

\subsection{Ablative Experiments}
\label{Ablations}

\begin{figure}[b!]
\centering
    \includegraphics[width=0.55\columnwidth]{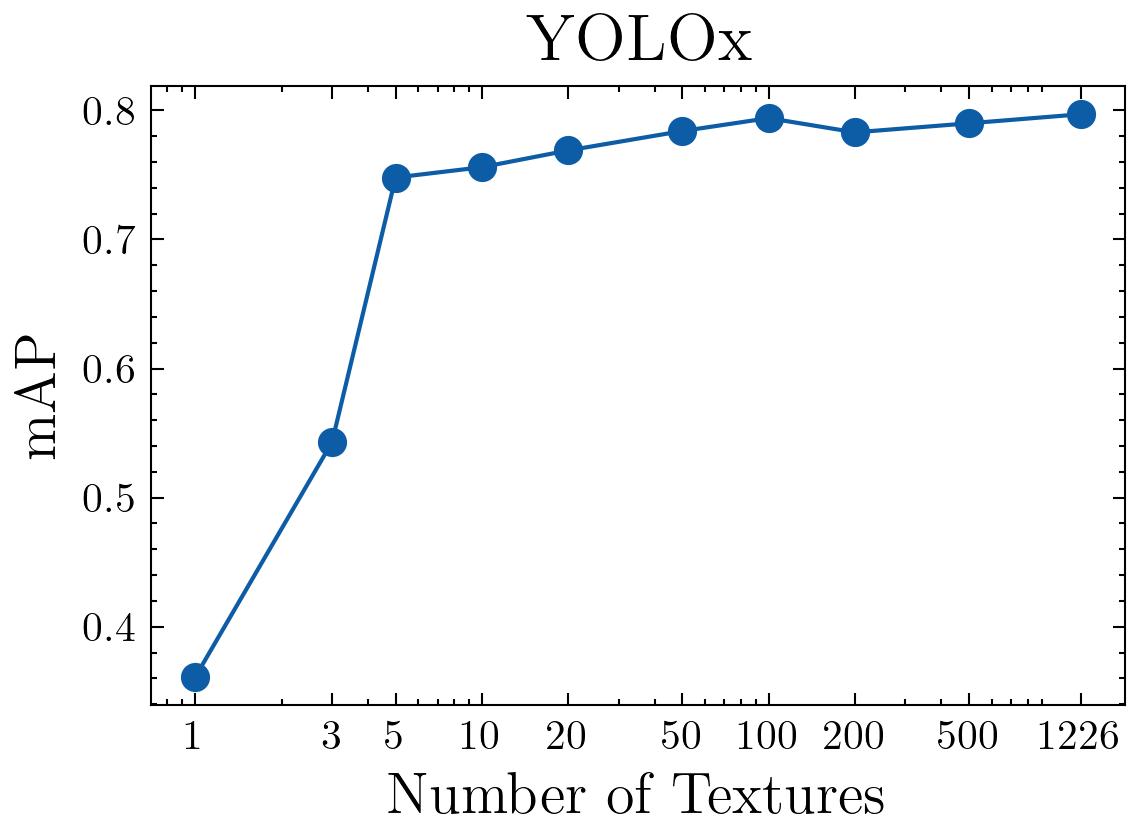}
\caption{\textbf{Number of Textures.} Ablation on the influence of the number of random textures for YOLOx on TLESS}
\label{fig:ablation_texture_count}
\end{figure}

This section presents ablative studies.
Reported are the influence of the number of random textures $n$, object detection and pose estimation at the superposition of our shape bias and online data augmentations, and the influence of the object geometry prior.

\subsubsection{Number of Random Textures}

Figure~\ref{fig:ablation_texture_count} shows the trend of YOLOx' mAP on TLESS when inducing a shape bias with a different number of random textures.
With the availability of more textures mAP improves.
Ultimately, using the $n=1226$ available cc-textures yields the highest precision.
However, using five random textures already results in $93.85\%$ and $100$ textures achieve $99.63\%$ of the maximally achieved pose estimation accuracy.

\begin{table}[b!]
\centering
\caption{\textbf{Data Augmentation.} Baselines for object detection and pose estimation; $p$ denotes augmentation probability}
\label{table:augmentation}
    \begin{tabular}{ccccc}
        \toprule
        \multicolumn{2}{c}{\textbf{Object Detection}} & & \multicolumn{2}{c}{\textbf{Pose Estimation}} \\
        \cmidrule{1-2}
        \cmidrule{4-5}        
        Augmentation & $p$ & & Augmentation & $p$\\
        \midrule 
        Coarse Dropout & $0.5$ & & Coarse Dropout & $0.5$ \\
        Gaussian Blur & $0.4$ & & Gaussian Blur & $0.5$ \\
        Enhance Sharpness & $0.3$ & & Add & $0.5$ \\
        Enhance Contrast & $0.3$ & & Invert & $0.3$ \\
        Enhance Brightness & $0.5$ & & Multiply & $0.5$ \\
        Enhance Color & $0.3$ & & Linear Contrast & $0.5$ \\
        Add & $0.5$ \\
        Invert & $0.3$ \\
        Multiply & $0.5$ \\
        Gaussian Noise & $0.1$ \\
        Linear Contrast & $0.5$ \\
        \bottomrule
    \end{tabular}
\end{table}

\begin{figure}[t!]
   \centering
    \includegraphics[width=1\columnwidth]{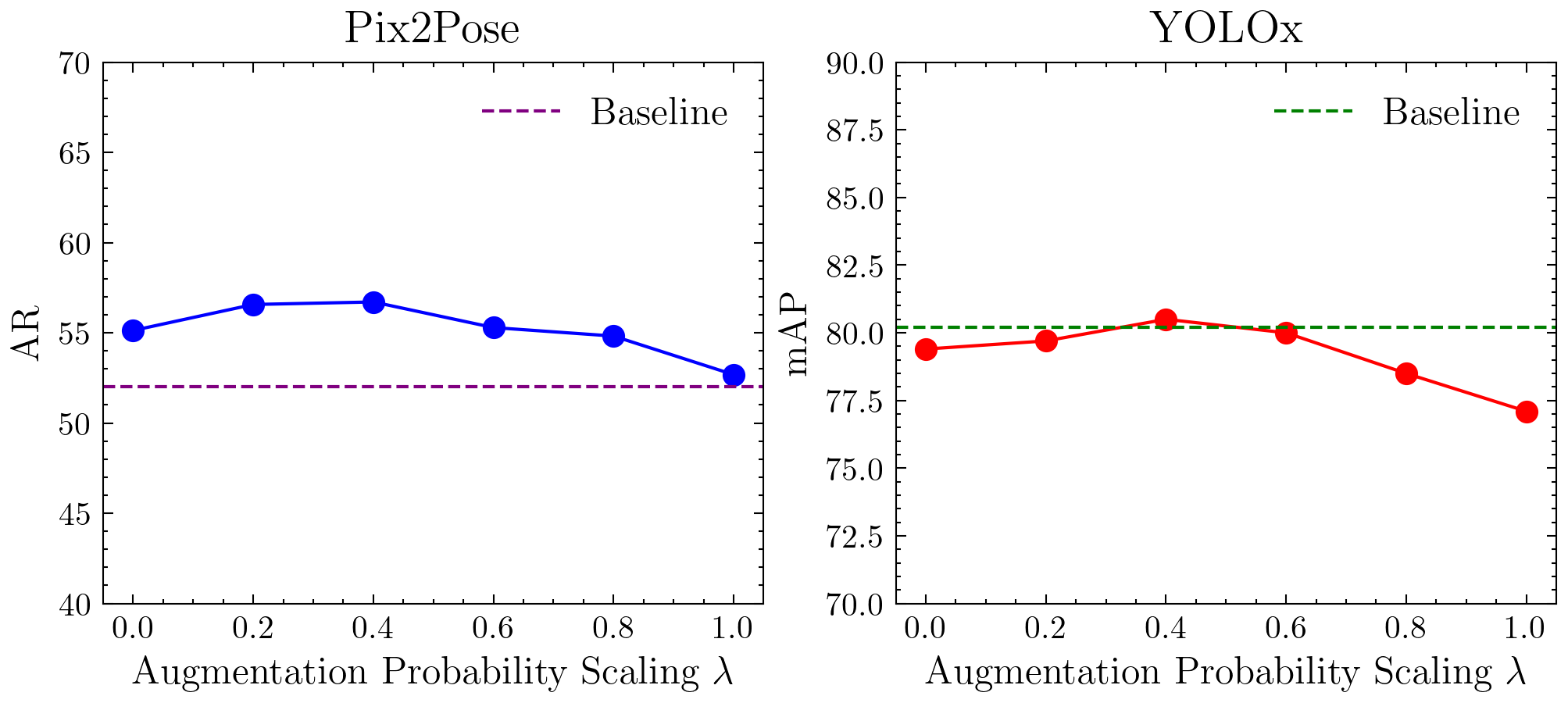}
   \caption{\textbf{Online Data Augmentation Severity.} Ablation study on the severity of online augmentation, 
 for Pix2Pose and YOLOx on TLESS; ground truth detection priors for Pix2Pose.}
   \label{fig:augmenation_prob_ablation}
\end{figure}

\subsubsection{Superposition of Shape Bias and Data Augmentation}
\label{sec:exp_dataaug}

This section provides a detailed analysis of the interplay of our shape bias with online data augmentation.
We perform a grid search over the augmentation probabilities used by~\cite{liu2022gdrnpp_bop} and~\cite{wang2021gdrnet}.
Table~\ref{table:augmentation} lists the types and probabilities of augmentations applied to object detection and pose estimation.
As done by GDR-Net~\cite{wang2021gdrnet}, and adopted for YOLOx by~\cite{liu2022gdrnpp_bop}, a coefficient $\lambda \in [0, 1]$ for scaling the probability of the applied data augmentations is defined.
The left part of Figure~\ref{fig:augmenation_prob_ablation} demonstrates results for object pose estimation with Pix2Pose, the right part shows that for detection with YOLOx on TLESS. 
For pose estimation a $\lambda$ of \(0.2 \leq \lambda \leq 0.4\) results in the most accurate estimates.
For detection a $\lambda$ of $0.4$ achieves the best results.
Generally, while shape-biased pose estimation improves with any set of online data augmentations, the data augmentation for object detection requires tuning to result in the highest mAP when training for a shape bias.

\subsubsection{Influence of the Mesh Origin}

This sections ablates the influence of the mesh origin and the type connected bias.
Table~\ref{table:ablation_reconstruction} reports the AR when training Pix2Pose for TLESS.
Ground truth detections are used as image location priors.
The results show that inducing a shape bias with reconstructed objects improves over using the object prior and its reconstructed texture.
The accurate geometry prior of CAD models is in general preferable over the reconstruction, yet again, training for a shape bias improves over training for a color bias; color bias refers to sampling object colors in grayscale, according to the parameters in Table~\ref{table:domainrandomization} and~\cite{sundermeyer2023bop}. 

\begin{table}[t!]
\centering
\caption{\textbf{Mesh Origin.} Pose estimation accuracy using Pix2Pose with diverse combination of object geometry and texture, on TLESS. The ground truth detections and an augmentation probability scaling factor of $\lambda=0.2$ are used.}
\label{table:ablation_reconstruction}
    \begin{tabular}{cc|c}
        \toprule
        
        \textbf{Mesh Origin} & \textbf{Bias} & \textbf{AR} \\
        \midrule
        Reconstructed & texture & 42.64 \\
        Reconstructed & \textbf{none (ours)} & 46.57 \\
        \cdashline{1-3}
        CAD & color & 59.24 \\
        CAD & \textbf{none (ours)} & \textbf{62.95} \\
        \bottomrule
    \end{tabular}
\end{table}

\section{Conclusion}
\label{Conclusion}
This paper presents a simple and easily applicable strategy for improving pose estimation of textureless and metallic objects.
Inducing a texture bias to estimators by randomizing object textures during data rendering results in negligible computational overhead to the standard strategy.
The improvements, however, are manifold.
Object detection improves in most of the tested cases, particularly for the most recent detector, pose estimation is on par or improves for both tested object detectors, the pose estimators are more robust to image perturbations.

Future studies will investigate adversarial learning for biasing estimators for shape, in order to further improve detection and pose estimation accuracy and robustness.
The presented study does not consider Vision Transformers and Diffusion models.
Future work will account for them.
Furthermore, since ITODD does not provide ground truth annotations, future work will provide more extensive evaluations for metallic objects.

\section*{Acknowledgement}

We gratefully acknowledge the support of the EU-program EC Horizon 2020 for Research and Innovation under grant agreement No. 101017089, project TraceBot, the Austrian Science Fund (FWF), under project No. I 6114, project iChores, and the city of Vienna (MA23 – Economic Affairs, Labour and Statistics) through the research project AIAV (MA23 project 26-04).

{\small
\bibliographystyle{ieee_fullname}
\bibliography{PaperForReview}

\begin{thebibliography}{10}\itemsep=-1pt

\bibitem{bauer2020verefine}
Dominik Bauer, Timothy Patten, and Markus Vincze.
\newblock {VeREFINE}: Integrating object pose verification with physics-guided iterative refinement.
\newblock {\em IEEE Robotics and Automation Letters}, 5(3):4289--4296, 2020.

\bibitem{behpour2019ada}
Sima Behpour, Kris~M. Kitani, and Brian~D. Ziebart.
\newblock Ada: Adversarial data augmentation for object detection.
\newblock In {\em Preceedings of the IEEE Winter Conference on Applications of Computer Vision}, pages 1243--1252, 2019.

\bibitem{denninger2019blenderproc}
Maximilian Denninger et~al.
\newblock Blenderproc.
\newblock {\em CoRR}, abs/1911.01911, 2019.

\bibitem{drost2017itodd}
Bertram Drost, Markus Ulrich, Paul Bergmann, Philipp Hartinger, and Carsten Steger.
\newblock Introducing {MVTec} {ITODD} - {A} dataset for {3D} object recognition in industry.
\newblock In {\em Proceedings of the IEEE International Conference on Computer Vision Workshops}, pages 2200--2208, 2017.

\bibitem{gatys2016image}
Leon~A Gatys, Alexander~S Ecker, and Matthias Bethge.
\newblock Image style transfer using convolutional neural networks.
\newblock In {\em Proceedings of the IEEE Conference on Computer Vision and Pattern Recognition}, pages 2414--2423, 2016.

\bibitem{ge2021yolox}
Zheng Ge, Songtao Liu, Feng Wang, Zeming Li, and Jian Sun.
\newblock {YOLOX}: Exceeding {YOLO} series in 2021.
\newblock {\em arXiv preprint arXiv:2107.08430}, 2021.

\bibitem{geirhos2018imagenettrained}
Robert Geirhos, Patricia Rubisch, Claudio Michaelis, Matthias Bethge, Felix~A. Wichmann, and Wieland Brendel.
\newblock Imagenet-trained {CNN}s are biased towards texture; increasing shape bias improves accuracy and robustness.
\newblock In {\em Proceedings of the International Conference on Learning Representations}, 2019.

\bibitem{geirhos2018generalisation}
Robert Geirhos, Carlos~RM Temme, Jonas Rauber, Heiko~H Sch{\"u}tt, Matthias Bethge, and Felix~A Wichmann.
\newblock Generalisation in humans and deep neural networks.
\newblock {\em Advances in Neural Information Processing Systems}, 31, 2018.

\bibitem{he2017mask}
Kaiming He, Georgia Gkioxari, Piotr Doll{\'a}r, and Ross Girshick.
\newblock Mask {R-CNN}.
\newblock In {\em Proceedings of the IEEE International Conference on Computer Vision}, pages 2961--2969, 2017.

\bibitem{he2016deep}
Kaiming He, Xiangyu Zhang, Shaoqing Ren, and Jian Sun.
\newblock Deep residual learning for image recognition.
\newblock In {\em Proceedings of the IEEE Conference on Computer Vision and Pattern Recognition}, pages 770--778, 2016.

\bibitem{he2023g}
Zaixing He, Yue Chao, Mengtian Wu, Yilong Hu, and Xinyue Zhao.
\newblock {G-GOP}: Generative pose estimation of reflective texture-less metal parts with global-observation-point priors.
\newblock {\em IEEE/ASME Transactions on Mechatronics}, 29(1):154--165, 2023.

\bibitem{zeixing2023contourpose}
Zaixing He et~al.
\newblock {ContourPose}: Monocular {6-D} pose estimation method for reflective textureless metal parts.
\newblock {\em IEEE Transactions on Robotics}, 39(5):4037--4050, 2023.

\bibitem{hodan2018bop}
Tom{\'a}{\v{s}} Hoda{\v{n}} et~al.
\newblock {BOP}: Benchmark for {6D} object pose estimation.
\newblock In {\em Proceedings of the European Conference on Computer Vision}, pages 19--34, 2018.

\bibitem{hodan2019photorealistic}
Tom{\'a}{\v{s}} Hoda{\v{n}} et~al.
\newblock Photorealistic image synthesis for object instance detection.
\newblock In {\em Proceedings of the IEEE International Conference on Image Processing}, pages 66--70. IEEE, 2019.

\bibitem{hodan2020bop}
Tom{\'a}{\v{s}} Hoda{\v{n}} et~al.
\newblock {BOP} challenge 2020 on {6D} object localization.
\newblock {\em Proceedings of the European Conference on Compututer Vision Workshops}, 2020.

\bibitem{hodan2017tless}
Tom{\'a}{\v{s}} Hoda{\v{n}}, Pavel Haluza, {\v{S}}tep{\'a}n Obdr{\v{z}}{\'a}lek, Jiri Matas, Manolis Lourakis, and Xenophon Zabulis.
\newblock {T-LESS}: An {RGB-D} dataset for {6D} pose estimation of texture-less objects.
\newblock In {\em Proceedings of the IEEE Winter Conference on Applications of Computer Vision}, pages 880--888, 2017.

\bibitem{hu2022perspective}
Yinlin Hu, Pascal Fua, and Mathieu Salzmann.
\newblock Perspective flow aggregation for data-limited 6d object pose estimation.
\newblock In {\em Proceedings of the European Conference on Computer Vision}, pages 89--106. Springer, 2022.

\bibitem{hu2021wide}
Yinlin Hu, Sebastien Speierer, Wenzel Jakob, Pascal Fua, and Mathieu Salzmann.
\newblock Wide-depth-range {6D} object pose estimation in space.
\newblock In {\em Proceedings of the IEEE/CVF Conference on Computer Vision and Pattern Recognition}, pages 15870--15879, 2021.

\bibitem{huang2017arbitrary}
Xun Huang and Serge Belongie.
\newblock Arbitrary style transfer in real-time with adaptive instance normalization.
\newblock In {\em Proceedings of the IEEE International Conference on Computer Vision}, pages 1501--1510, 2017.

\bibitem{kaskman2019homebreweddb}
Roman Kaskman, Sergey Zakharov, Ivan Shugurov, and Slobodan Ilic.
\newblock {HomebrewedDB}: {RGB-D} dataset for {6D} pose estimation of {3D} objects.
\newblock In {\em Proceedings of the IEEE/CVF International Conference on Computer Vision Workshops}, 2019.

\bibitem{labbe2020cosypose}
Yann Labb{\'e}, Justin Carpentier, Mathieu Aubry, and Josef Sivic.
\newblock Cosypose: Consistent multi-view multi-object 6d pose estimation.
\newblock In {\em Proceedings of the European Conference on Computer Vision}, pages 574--591. Springer, 2020.

\bibitem{lin2014microsoft}
Tsung-Yi Lin et~al.
\newblock Microsoft {COCO}: Common objects in context.
\newblock In {\em Proceedings of the European Conference on Computer Vision}, pages 740--755, 2014.

\bibitem{lin2017focal}
Tsung-Yi Lin, Priya Goyal, Ross Girshick, Kaiming He, and Piotr Doll{\'a}r.
\newblock Focal loss for dense object detection.
\newblock In {\em Proceedings of the IEEE International Conference on Computer Vision}, pages 2980--2988, 2017.

\bibitem{liu2022gdrnpp_bop}
Xingyu Liu, Ruida Zhang, Chenyangguang Zhang, Bowen Fu, Jiwen Tang, Xiquan Liang, Jingyi Tang, Xiaotian Cheng, Yukang Zhang, Gu Wang, and Xiangyang Ji.
\newblock Gdrnpp.
\newblock \url{https://github.com/shanice-l/gdrnpp_bop2022}, 2022.

\bibitem{mildenhall2021nerf}
Ben Mildenhall, Pratul~P Srinivasan, Matthew Tancik, Jonathan~T Barron, Ravi Ramamoorthi, and Ren Ng.
\newblock {NeRF}: Representing scenes as neural radiance fields for view synthesis.
\newblock {\em Communications of the ACM}, 65(1):99--106, 2021.

\bibitem{nie2020total3dunderstanding}
Yinyu Nie, Xiaoguang Han, Shihui Guo, Yujian Zheng, Jian Chang, and Jian~Jun Zhang.
\newblock {Total3DUnderstanding}: Joint layout, object pose and mesh reconstruction for indoor scenes from a single image.
\newblock In {\em Proceedings of the IEEE/CVF Conference on Computer Vision and Pattern Recognition}, pages 55--64, 2020.

\bibitem{park2019pix2pose}
Kiru Park, Timothy Patten, and Markus Vincze.
\newblock {Pix2Pose}: Pixel-wise coordinate regression of objects for {6D} pose estimation.
\newblock In {\em Proceedings of the IEEE/CVF International Conference on Computer Vision}, Oct 2019.

\bibitem{ren2015faster}
Shaoqing Ren, Kaiming He, Ross Girshick, and Jian Sun.
\newblock Faster {R-CNN}: Towards real-time object detection with region proposal networks.
\newblock {\em Advances in Neural Information Processing Systems}, 28, 2015.

\bibitem{russakovsky2015ImageNet}
Olga Russakovsky, Jia Deng, Hao Su, Jonathan Krause, Sanjeev Satheesh, Sean Ma, Zhiheng Huang, Andrej Karpathy, Aditya Khosla, Michael Bernstein, Alexander~C. Berg, and Li Fei-Fei.
\newblock {ImageNet Large Scale Visual Recognition Challenge}.
\newblock {\em International Journal of Computer Vision}, 115(3):211--252, 2015.

\bibitem{sundermeyer2023bop}
Martin Sundermeyer et~al.
\newblock Bop challenge 2022 on detection, segmentation and pose estimation of specific rigid objects.
\newblock {\em Proceedings of the IEEE/CVF Conference on Computer Vision and Pattern Recognition}, 2023.

\bibitem{thalhammer2023challenges}
Stefan Thalhammer, Dominik Bauer, Peter H{\"o}nig, Jean-Baptiste Weibel, Jos{\'e} Garc{\'\i}a-Rodr{\'\i}guez, and Markus Vincze.
\newblock Challenges for monocular {6D} object pose estimation in robotics.
\newblock {\em arXiv preprint arXiv:2307.12172}, 2023.

\bibitem{thalhammer2021pyrapose}
Stefan Thalhammer, Markus Leitner, Timothy Patten, and Markus Vincze.
\newblock {PyraPose}: Feature pyramids for fast and accurate object pose estimation under domain shift.
\newblock In {\em IEEE International Conference on Robotics and Automation}, pages 13909--13915, 2021.

\bibitem{thalhammer2023cope}
Stefan Thalhammer, Timothy Patten, and Markus Vincze.
\newblock {COPE}: End-to-end trainable constant runtime object pose estimation.
\newblock In {\em Proceedings of the IEEE/CVF Winter Conference on Applications of Computer Vision}, pages 2859--2869, 2023.

\bibitem{tobin2017domainrandomization}
Josh Tobin, Rachel Fong, Alex Ray, Jonas Schneider, Wojciech Zaremba, and Pieter Abbeel.
\newblock Domain randomization for transferring deep neural networks from simulation to the real world.
\newblock In {\em Proceedings of the IEEE/RSJ International Conference on Intelligent Robots and Systems}, pages 23--30, 2017.

\bibitem{wang2021gdrnet}
Gu Wang, Fabian Manhardt, Federico Tombari, and Xiangyang Ji.
\newblock {GDR-Net}: Geometry-guided direct regression network for monocular {6D} object pose estimation.
\newblock In {\em Proceedings of the IEEE/CVF Conference on Computer Vision and Pattern Recognition}, pages 16611--16621, 2021.

\bibitem{jun2023texturelesspose}
Jun Yang, Wenjie Xue, Sahar Ghavidel, and Steven~L. Waslander.
\newblock {6D} pose estimation for textureless objects on {RGB} frames using multi-view optimization.
\newblock In {\em Proceedings of the IEEE International Conference on Robotics and Automation}, pages 2905--2912, 2023.

\bibitem{yang2018tsdf}
Long Yang, Qingan Yan, Yanping Fu, and Chunxia Xiao.
\newblock Surface reconstruction via fusing sparse-sequence of depth images.
\newblock {\em IEEE Transactions on Visualization and Computer Graphics}, 24(2):1190--1203, 2018.

\bibitem{yu2023tgf}
Haixin Yu, Shoujie Li, Houde Liu, Chongkun Xia, Wenbo Ding, and Bin Liang.
\newblock {TGF-Net}: {Sim2Real} transparent object {6D} pose estimation based on geometric fusion.
\newblock {\em IEEE Robotics and Automation Letters}, 8(6):3868--3875, 2023.

\bibitem{zoph2020learning}
Barret Zoph, Ekin~D Cubuk, Golnaz Ghiasi, Tsung-Yi Lin, Jonathon Shlens, and Quoc~V Le.
\newblock Learning data augmentation strategies for object detection.
\newblock In {\em Proceedings of the European Conference on Computer Vision}, pages 566--583. Springer, 2020.

\end{thebibliography}
}

\end{document}